\documentclass[10pt,twocolumn,letterpaper]{article}

\usepackage{cvpr}
\usepackage{times}
\usepackage{epsfig}
\usepackage{graphicx}
\usepackage{amsmath}
\usepackage{amssymb}

\usepackage{subfigure}
\usepackage{amsthm}

\usepackage{multirow}
\usepackage{booktabs}
\usepackage{algorithm}
\usepackage{algorithmic}
\usepackage{epstopdf}
\usepackage{authblk}

\newtheorem{theorem}{Theorem}
\newtheorem{proposition}[theorem]{Proposition}
\usepackage[pagebackref=true,breaklinks=true,letterpaper=true,colorlinks,bookmarks=false]{hyperref}

 \cvprfinalcopy % *** Uncomment this line for the final submission

\def\cvprPaperID{812} % *** Enter the CVPR Paper ID here

\ifcvprfinal\fi
\begin{document}

%%%%%%%%% TITLE
\title{Improved Search in Hamming Space using Deep Multi-Index Hashing}

\author[$^{\dagger}$]{Hanjiang Lai}
\author[$^{\ddagger}$]{Yan Pan \thanks{Corresponding author: Yan Pan, email: panyan5@mail.sysu.edu.cn.}}
\affil[$^{\ddagger}$]{School of Data and Computer Science, Sun Yan-Sen University, China}

%\author{Hanjiang Lai$^{\dagger\ddagger}$\and Yan Pan\thanks{Corresponding author: Yan Pan, email: panyan5@mail.sysu.edu.cn.}
% \and Canyi Lu$^{\dagger}$ \and Yong Tang$^\S$   \and Shuicheng Yan$^\dagger$}
%%Hefei Institute of Intelligent Machines, Chinese Academy of Sciences, Hefei, China,
%\institute{$^\dagger$ Department of Electrical and Computer Engineering, National University of Singapore \\
%$^\star$ School of Software, Sun Yat-sen University, China \\
%$^\ddagger$ School of Information Science and Technology, Sun Yat-sen University, China \\
%$^\S$ School of Computer Science, South China Normal University, China\\
%{\tt\small laihanj@gmail.com}  ~~
%{\tt\small panyan5@mail.sysu.edu.cn}  ~~ \\
%{\tt\small canyilu@gmail.com} ~~
%{\tt\small ytang@scnu.edu.cn} ~~
%{\tt\small eleyans@nus.edu.sg}
%}

\maketitle
%\thispagestyle{empty}

%%%%%%%%% ABSTRACT
\begin{abstract}
Similarity-preserving hashing is a widely-used method for nearest neighbour search in large-scale image retrieval tasks. There has been considerable research on generating efficient image representation via the deep-network-based hashing methods. However, the issue of efficient searching in the deep representation space remains largely unsolved. To this end, we propose a simple yet efficient deep-network-based multi-index hashing method for simultaneously learning the powerful image representation and the efficient searching. To achieve these two goals, we introduce the multi-index hashing (MIH) mechanism into the proposed deep architecture, which divides the binary codes into multiple substrings. Due to the non-uniformly distributed codes will result in inefficiency searching, we add the two balanced constraints at feature-level and instance-level, respectively. Extensive evaluations on several benchmark image retrieval datasets show that the learned balanced binary codes bring dramatic speedups and achieve comparable performance over the existing baselines. 
\end{abstract}

%%%%%%%%% BODY TEXT
\section{Introduction}
Nearest neighbor (NN) search has attracted increasing interest due to the ever-growing large-scale data on the web, which is a fundamental requirement in image retrieval~\cite{LSH}. Recently, similarity-preserving hashing methods that encode images into binary codes have been widely studied. Learning good hash functions should require two principles: 1) powerful image representation and 2) efficient searching in representation space. In this paper, we focus on deep-network-based hashing for efficient searching and keep the good performance.

In recent year, we have witnessed great success of deep neural networks, which the success mainly comes from the powerful representation learned from the deep network architectures. The deep-networks-based hashing methods learn the image representations as well as the binary hash codes. Lin et al.~\cite{lin2015deep} proposed a method that learns hash codes and image representations in a point-wised manner. Li et al.~\cite{li_pairwise} proposed a novel deep hashing method called deep pairwise-supervised hashing (DPSH) to perform simultaneous hash code learning and feature learning. Zhao et al.~\cite{zhao2015deep} presented a deep semantic ranking based method for learning hash functions that preserve multilevel semantic similarity for multi-label images. Further, Zhuang~\cite{zhuang2016fast} proposed a fast deep network for triplet supervised hashing.
 
Although the powerful binary codes have been learned from the deep networks, linear scan of Hamming distance is also time consuming in front of large-scale dataset (e.g., millions or billions images). Many methods have been proposed for efficient searching in Hamming space. One popular approach is to use binary codes as the indices into a hash table~\cite{semantic}. The problem is that the number of buckets grows near-exponentially. Norouzi et al.~\cite{mih} proposed multi-index hashing (MIH) method for fast searching, which divides the binary codes into smaller substrings and build multiple hash tables. MIH assumes that binary codes are uniformly distributed over the Hamming space, which is always not true. Liu et al.~\cite{liu2015data} and Wan et al.~\cite{wan2013data} proposed data-oriented multi-index hashing, respectively. They firstly calculated the correlation matrix between bits and then rearranged the indices of bits to make a more uniform distribution in each hash table. Ong et al.~\cite{ong2016improved} relaxed the equal-size constraint in MIH and proposed multiple hash tables with variable length hash keys. Wang et al.~\cite{wang2015multi} used repeat-bits in Hamming space to accelerate the searching but need more storage space. Song et al.~\cite{wangjd} proposed a distance-computation-free search scheme for hashing.

Most of the existing works firstly use the hashing models (e.g., LSH~\cite{LSH}, MLH~\cite{MLH}) to encode the image into the binary codes, followed by separate methods to rebalance the binary codes distribution. Such fixed hashing models may result in suboptimal searching. Ideally, it is expected that hash models and balanced procedure can be learned simultaneously during the hash learning process.

\begin{figure*}
  \centering
  \includegraphics[width=1\hsize]{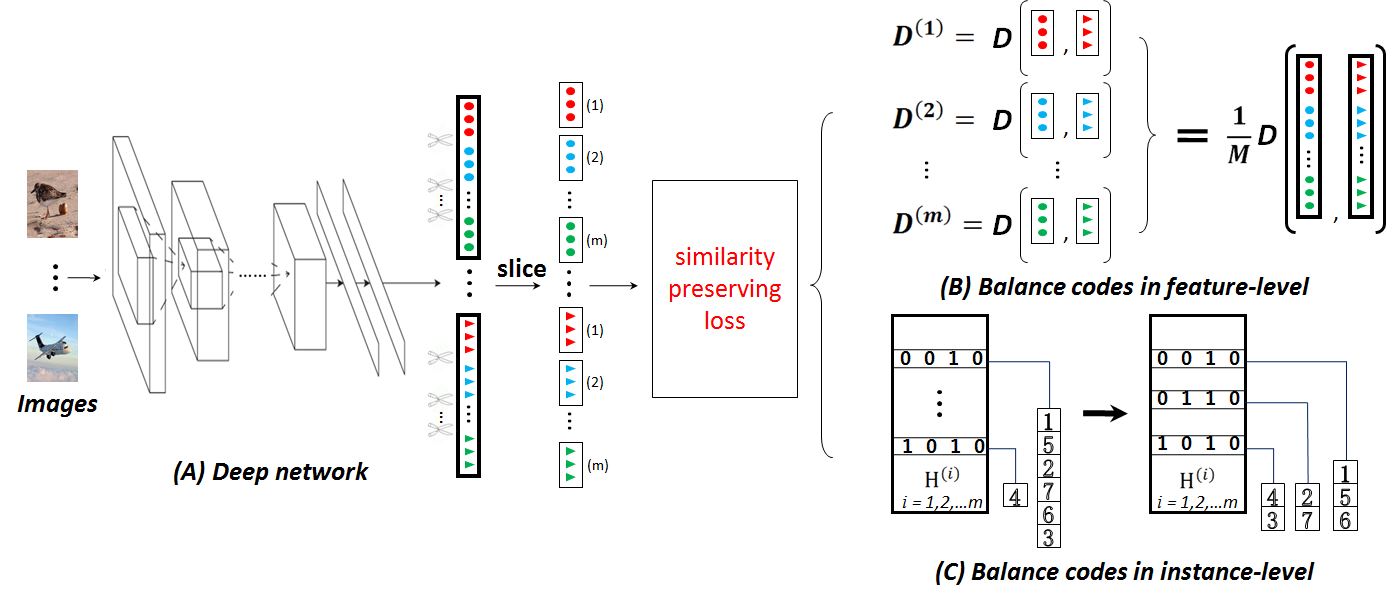}
  \caption{An overview of the proposed deep multi-index hashing architecture. It firstly (A) encodes the images into the image representations via stacked convolutional and fully-connected layers. Then the intermediate features are divided into $m$ slices with equal length. After that, we propose a similarity-preserving objective with two constraints for learning more uniformly distributed binary codes at feature-level and instance-level, respectively. The feature-level constraint (B) requires that there is an equal distance in each substring for any two binary codes, and the instance-level (C) lets the hash table buckets contain balanced items. }
    % be used to cite the whole fig
    \label{overview}
\end{figure*}

In this paper, we propose a deep architecture for fast searching and efficient image representation by incorporating the MIH approach into the network. As shown in Figure.~\ref{overview}, our architecture consists of three main building blocks. The first block is for learning the good image representation by the stacked convolutional and fully-connected layers followed by a slice layer which divides intermediate image features into multiple substrings, each substring corresponding to one hash table as the MIH approach. And the second and third blocks learn the uniform codes distribution, which balances the binary codes in feature-level and instance-level, respectively. In feature-level, we make the bits distributed as uniform as possible in each substring hash table by adding a new balanced constraint in the objective. And the instance-level is used to punish the buckets contain too many items, which will cost much time for checking many candidate codes. Finally, a similarity-preserving objective with two balanced constraints is proposed to capture the similarities among the images, and a fast hash model is learned to encode all the images into more uniformed binary codes.

The main contributions of this work are two-folds.
\begin{itemize}
\item We propose a deep multi-index hashing, which learns the hash functions for both the powerful image representation and fast searching.

\item We conduct extensive evaluations on several benchmark datasets. The empirical results demonstrate the superiority of the proposed method over the state-of-the-art baseline methods. 

\end{itemize}

\section{Related Work}

The learning-to-hash methods learn the hash functions from the training data for generating better binary representation. The representative methods include Iterative Quantization (ITQ)~\cite{ITQ}, Kernerlized LSH (KLSH) \cite{KLSH}, Anchor Graph Hashing (AGH)~\cite{AGH}, Spectral Hashing (SH)~\cite{sh}, Semi-Supervised Hashing (SSH)~\cite{SSH}, Kernel-based Supervised Hashing (KSH)~\cite{KSH}, Minimal Loss Hashing (MIH)~\cite{MLH}, Binary Reconstruction Embedding (BRE)~\cite{BRE} and so on. The comprehensive survey can be found in~\cite{wangjd2}.

Deep-network-based hashing method has been emerged as one of the leading approaches. Many algorithms ~\cite{jiang2016deep,nlabel,liu2017deep,zhang2016efficient,zhuang2016fast,lin2015deep,yang2015supervised,zhang2015bit,zhao2015deep} have been proposed, including the point-wise approach, the pair-wise approach and the ranking-based approach. The point-wise methods take a single image as input and the loss function is built on individual data. For example, Lin et al.~\cite{lin2015deep} showed that the binary codes can be learned by employing a hidden layer for representing the latent concepts that dominate the class labels, thus they proposed to learn the hash codes and image representations in a point-wised manner. Yang et al.~\cite{yang2015supervised} proposed a loss function defined on classification error and other desirable properties of hash codes to learn the hash functions. The pair-wise methods take the image pairs as input and the loss functions are used to characterize the relationship (i.e., similar or dissimilar) between a pair of two images. Specifically, if two images are similar, then the hamming distance between the two images should be small, otherwise, the distance should be large. Representative methods include deep pairwise-supervised hashing (DPSH)~\cite{li_pairwise}, deep supervised hashing (DSH)~\cite{liu2016deep} and so on. The ranking-based methods cast learning-to-hash as the ranking problem. % Lai et al.~\cite{onestep} proposed a one-stage supervised hashing method which takes three images as input and preserve relative similarities of the form ``image $I$ is more similar to image $I^+$ than to image $I^-$''.
Zhao et al.~\cite{zhao2015deep} proposed a deep semantic ranking-based method for learning hash functions that preserve multi-level semantic similarity between multi-label images. Zhuang~\cite{zhuang2016fast} proposed a fast deep network for triplet supervised hashing. 
%The ranking-based methods can be applied in both the single-label and multi-label image retrieval. 

Although obtaining the powerful image representation via the deep learning-to-hash methods, existing works always do not consider the fast searching in the learned codes space.
% Linear scan can be used for finding near neighbours, which adopts the exhaustive search scheme to calculate the Hamming distances between the query and all data in the database. While, it is prohibitive when searching in a large-scale database.
Multi-index hashing~\cite{greene1994multi,mih} is an efficient method for finding all $r$-neighbors of a query by dividing the binary codes into multiple substrings. While, binary codes learned from the deep network always not be uniformly distributed in practice, e.g., all images with the same label indices with a similar key as shown in Figure~\ref{example}, which will cost much time to check many candidate codes. In this paper, we solve this problem by adding two balanced constraints in our network, and learn more uniformly distributed binary codes.

\section{Background: Multi-Index Hashing}
In this section, we briefly review MIH~\cite{mih}.

MIH is a method for fast searching in large-scale datasets, which the binary code $ \mathbf{h}$ is partitioned into $m$ disjoint substring, $\mathbf{h}^{(1)},\cdots,\mathbf{h}^{(m)}$, each substring consists of $l/m $ bits, where $l$ is the length of bits and we assume $l$ is divisible by $m$ for convenience presentation. One hash table is built for each of the $m$ substrings. 

The $r$-\textit{neighbor} of a query $\mathbf{q}$ is denoted as $\mathcal{R}^{r}(\mathbf{q})$ which differ from $\mathbf{q}$ in $r$ bits or less from all codes in the database. 
%The $r$\textit{-neighbor search problem} is defined as: find all $r$-\textit{neighbor} of a query  $\mathbf{q}$. 
To search the $r$-neighbor of a query $\mathbf{q}$ with substrings $\{\mathbf{q}^{(j)}\}_{j=1}^m$, MIH searches the all substring hash tables for entries that are within a Hamming distance of $r' = r/m$~\footnote{For ease of presentation, here we assume $r$ is divisible by $m$. In practice, if $r=m\times r'+a$ with $0<a<m$, we can set the search radii of the first $a+1$ hash tables to be $r'$ and the rests to be $r'-1$.}. The set of candidates from the $j$-th substring hash table is denoted as $\mathcal{N}_j^{r'}(\mathbf{q})$. Then, the union of all the $m$ sets, $\mathcal{N}^{r'}(\mathbf{q})= \bigcup_{j=1}^{m} \mathcal{N}_j^{r'}(\mathbf{q})$, is the superset of the $r$-neighbors of $\mathbf{q}$. The false positives that are not true $r$-neighbors of $\mathbf{q}$ are removed by computing the full Hamming distance.

The $k$NN \textit{search problem} can be formulated as the $r$-near neighbor problem. By initializing integer $r'=0$, we can progressive increment of the search radius $r'$ until the specified number of neighbors is found.

\section{Deep Multi-Index Hashing}
This section describes deep multi-index hashing architecture that allows us to 1) obtain powerful binary codes and 2) efficient search inside the binary codes space.  

We firstly introduce notations. There is a labeled training set $\{I_i, y_i\}_{i=1}^{n}$, where $I_i$ is the $i$-th image, $y_i$ is the class name/label of the $i$-th image, and the number of training samples is $n$. Suppose that each binary code comprises $l$ bits, the goal of deep multi-index hashing is to learn a deep hash model, in which the similarities among the binary codes should be preserved and also quick searching in large-scale binary codes space. 

As shown in Figure~\ref{overview}, the purposes of the proposed architecture are two: 1) a deep network with multiple convolution-pooling layers to capture an efficient representation of images, and followed by a slice layer to partition the feature into $m$ disjoint substrings, and 2) a balanced codes module designed to address the ability to quickly search inside the binary codes space. It generates the binary codes distributed as uniform as possible in each substring hash table from two aspects: feature-level and instance-level. In the following, we will present the details of these parts, respectively.

\begin{figure}
  \centering
  \includegraphics[width=0.9\hsize]{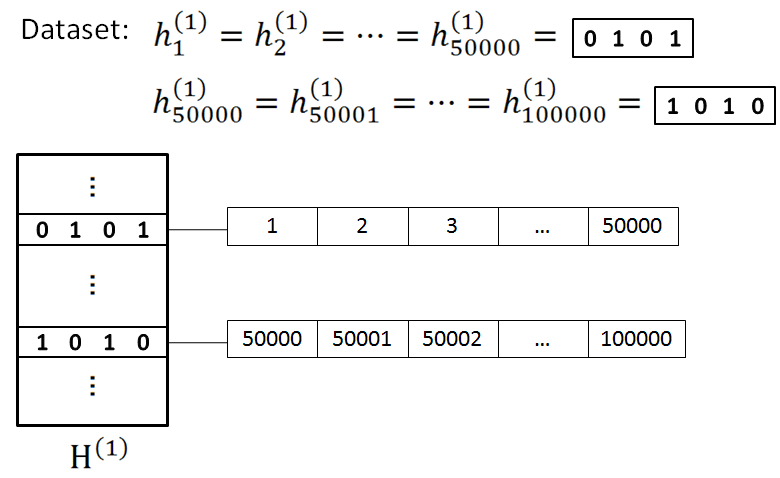}
  \caption{An example of powerful binary codes while bad searching in codes space.  }
    % be used to cite the whole fig
    \label{example}
\end{figure}

\subsection{Efficient Representation via Deep Network}
The deep network, e.g., AlexNet~\cite{AlexNet}, VGG~\cite{VGG}, GoogleLeNet~\cite{GoogleLeNet} and residual network~\cite{he2016deep}, is used for learning the powerful efficient image representation, which is made following structural modifications for image retrieval task. The first modification is to remove the last fully-connected layer (e.g., fc8). The second is to add a fully-connected layer with $l$ dimensional intermediate features. The intermediate features are then fed into a tanh layer that restricts the values in the range $[-1,1]$. The MIH contains $m$ separate hash tables. Inspired by that, the third modification is to add a slice layer to divide the features into $m$ slices with equal length $l/m$. According to the suggestion of the MIH, the number of substring hash  tables is setted to be $m = \lfloor \frac{l}{\log_2 n} \rfloor$, which shows the best empirical performance shown in~\cite{mih}. Finally, the output of network is denoted as $\mathbf{h}_i = \mathcal{F}(I_i), \mathbf{h}_i = [\mathbf{h}_i^{(1)},\cdots,\mathbf{h}_i^{(m)}]$, where $I_i$ is the input image and $\mathcal{F}$ is the deep network.

%With the proposed network with stacked layers, we can use pairwise loss that indicates the semantic similarities/dissimilarities on image pairs:
%\begin{equation}
%\begin{aligned}
%& \min \sum_{i,j} S_{ij} D(\mathbf{h}_i,\mathbf{h}_j) + (1-S_{ij}) \max(0, \epsilon - D(\mathbf{h}_i,\mathbf{h}_j)),& \\
%\end{aligned}
%\end{equation}
%or triple loss that preserves relative similarities of the form ``$\mathbf{h}_i$ is more similar to $\mathbf{h}_{i^+}$ than to $\mathbf{h}_{i^-}$'':
%\begin{equation}
%\begin{aligned}
%& \min \sum_{i,i^+,i^-} \max(0, \epsilon + D(\mathbf{h}_i,\mathbf{h}_{i^+}) -  D(\mathbf{h}_i,\mathbf{h}_{i^-})),& \\
%\end{aligned}
%\end{equation}
%to learn the similarity-preserving hash functions, where $\epsilon$ is a parameter, $D(\mathbf{h}_i,\mathbf{h}_j)$ is the distance between two binary codes, e.g., the euclidean distance or the cosine similarity, and $S_{ij}=1$ if the two images are similar, otherwise $S_{ij} = 0$. 

The deep-network based methods can learn a very powerful image representation for the image retrieval, while they do not consider the ability to efficiently search inside the representation space. An example of powerful image representation for binary codes while bad searching is shown in Figure~\ref{example}. Here the substring of length $4$. Suppose that there are 2 class labels, and each class consists of 50,000 images. Without loss of generality, the first 50,000 images whose labels are $1$, the labels of the rest 50,000 images are $2$. The hash table is built for the 100,000 learned binary codes shown in Figure~\ref{example}, where the similar codes locate in the same bucket (with a similar key) and the dissimilar codes have largest Hamming distance, i.e., $4$, in the hash table. The learned binary codes are very good for accuracy while they are very bad for searching. Given a query, it needs to check so many candidate items (e.g., 50,000 items). It is necessary for finding a new way to generate more balanced binary codes.  

% the corresponding similarity matrix is shown in Figure~\ref{block_diagonal} (a). In this matrix, there are three block-diagonals (e.g., the first two columns are identical). It is obvious that the similarity matrix has a block-diagonal structure and the rank of the matrix is 3.  In general, for any similarity matrix generated by $c$ class labels, it can be converted to a block-diagonal matrix with $c$ blocks. The rank of the matrix should be $c$. 

%with $1\times 1$ filters after some convolution layers with filters of a larger receptive field. These $1\times 1$ convolution filters can be regarded as a linear transformation of their input channels (followed by rectification non-linearity). As suggested in~\cite{NIN}, we use an average-pooling layer as the output layer of this sub-network, to replace the fully-connected layer(s) used in traditional architectures (e.g.,~\cite{AlexNet}). As an example, Table \ref{NIN_def} shows the configurations of the sub-network for images of size $256 \times 256$.

%We propose to use a deep network with a stack of convolution layers to automatically learn a powerful representation of the input images. Through this network, an image is encoded to a $l$-dimensional feature. Then the feature is divided into $m$ slices with equal length $l/m$\footnote{For ease of presentation, here we assume the dimension $l$ is divisible by $m$. In practice, if $l=m\times s +c$ with $0<c<m$, we can set the first $c$ slices to be length of $s+1$ and the rest $m-c$ ones to be length of $s$.}. 

\subsection{Fast Searching via the Deep Multi-Index Hashing}
% This leads to the following proposition. 

%The MIH contains $m$ separate hashtables. Inspired by that, we add a slice layer to divide the features into $m$ slices with equal length $l/m$. According to the suggestion of the MIH, the number of substring hash tables is setted to be $m = \lfloor \frac{l}{\log_2 n} \rfloor$, which shows the best empirical performance shown in~\cite{mih}. 

%Formally, given a query $q$ with substring $\{q^{(i)}\}_{i=1}^m$, we search the $i$-th substring hash table for items that are within a Hamming distance of $r'$. The set of retrieved samples from the $i$-th substring table is denoted as $\mathcal{N}_i^{r'}(q)$. We also denote that $r$-neighbor of a query $q$ as $\mathcal{R}^{r}(q)$ which differ from $q$ in $r$ bits or less from all codes in the database.   

We first give the following proposition.
\begin{proposition}
When the buckets in the substring hash tables that differ from $\mathbf{q}^{(j)}$ within $r'$ bits, i.e., $D(\mathbf{h}^{(j)}, \mathbf{q}^{(j)}) \leq r'$, then we have $\mathcal{R}^{r}(\mathbf{q}) \subseteq \bigcup_{i=1}^{j} \mathcal{N}_i^{r'}(\mathbf{q})$, where $r = r'm + j - 1$.
\label{proposition1}
\end{proposition}

For example, suppose that $r'=0$, when searching in the first substring hash table, we obtain a set of candidates $\mathcal{N}_1^{0}(\mathbf{q})$, then the $0$-neighbor (i.e., $r=r'm +i -1 = 0$) of query $\mathbf{q}$ is the subset of the candidates, that is  $\mathcal{R}^{0}(\mathbf{q}) \subseteq \mathcal{N}_1^{0}(\mathbf{q})$. Similar, we have $\mathcal{R}^{1}(\mathbf{q}) \subseteq \bigcup_{i=1}^{2} \mathcal{N}_i^{0}(\mathbf{q})$, $\mathcal{R}^{2}(\mathbf{q}) \subseteq \bigcup_{i=1}^{3} \mathcal{N}_i^{0}(\mathbf{q})$ and etc. When searching in the substring hash tables differs by $r'$ bits or less, we can obtain all $r$-neighbor of the query, where $r'm \leq r < (r'+1)m$.

According to the above proposition, we can see that the running time of MIH for $k$NN mainly contains two parts: index lookups and candidate codes checking. To achieve faster searching, we should reduce 1) the number of distinct hash buckets to examine, i.e., the smaller $r'$, the better. 2) the number of candidate codes to check, i.e., the smaller $\bigcup_{j=1}^{i} \mathcal{N}_j^{r'}(\mathbf{q})-\mathcal{R}^{r}(\mathbf{q})$, the better.

\subsubsection{Balanced binary codes in Instance-level} To reduce the running time for index lookups, the binary codes of similar images should be indices with a similar key as shown in Figure~\ref{example}. In such case, $r'=0$. Unfortunately, we need to check so many candidate binary codes, making the inefficient searching. Thus, the number of each bucket should  be not too small and not too large. Balanced binary codes in instance-level are learned for addressing the problem, which require that each bucket in the hash table contains at most $k$ items.

\begin{figure}
  \centering
  \includegraphics[width=1\hsize]{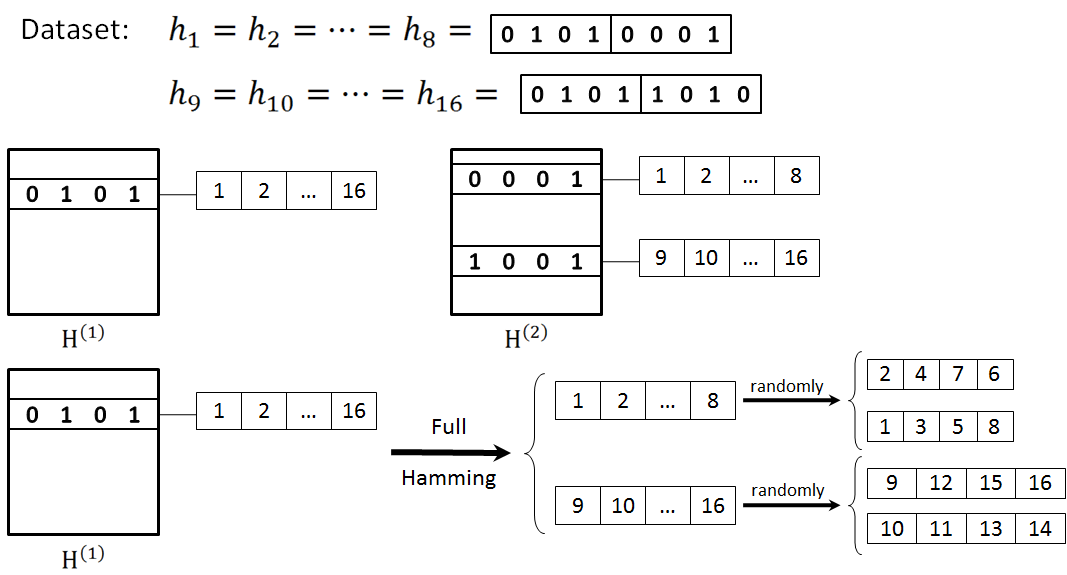}
  \caption{An example for rebalancing the items in the bucket. Here the codes is split into 2 hash tables $H^{(1)}$ and $H^{(2)}$. We require each bucket contains at most $4$ items. Since the haskey $(0101)$ in hash table $H^{(1)}$ has 16 items, we first use full Hamming distance to split the items into 2 groups. Then the items in groups are further randomly splitted into subgroups until all subgroups contain not more than $k'$ items.  }
    % be used to cite the whole fig
    \label{example2}
\end{figure}

Formally, all the buckets in all substring hash tables that have more than $k'$ items were found. Let $P^{(j)}_i=\{p_1,\cdots,p_{c}\}$ is denoted as the items in the $i$-th bucket of the $j$-th substring hash table. We use the following steps to rebalance these items as shown in Figure~\ref{example2}.

1) The full Hamming distance is used to split these items into several groups, each group contains the samples which have the same binary codes. If the number of all groups are less than $k'$, stop the procedure. 

2) Otherwise, if the number of the group is more than $k'$, we further randomly split it into $n_g/k'$ subgroups with the equal sizes, making sure each subgroup consists of at most $k'$ items, where $n_g$ is the number of items in the group.

A key principle should be ensured is that do not change the similarities among these images, that is the distance between $p_a$ and $p_b$, i.e., $D(\mathbf{h}_{p_a}^{(j)},\mathbf{h}_{p_b}^{(j)})$, should preserve relative similarities of the form ``($p_a,p_b$ in the same subgroup) $\leq$ ($p_a,p_b$ in the same group) $\leq$ ($p_a,p_b$ in the different groups)". Thus, the objective can be formulated as: $\phi(P^{(j)}(i)) = $
\begin{displaymath}
 \left\{ \begin{array}{ll}
D(\mathbf{h}_{p_a}^{(j)},\mathbf{h}_{p_b}^{(j)}) & \textrm{if
$p_a,p_b$ in  same subgroup} \\
\max(0, 1 - D(\mathbf{h}_{p_a}^{(j)},\mathbf{h}_{p_b}^{(j)})) & \textrm{if $p_a,p_b$ in  same group} \\
\max(0, 2 - D(\mathbf{h}_{p_a}^{(j)},\mathbf{h}_{p_b}^{(j)})) & \textrm{if $p_a,p_b$ in different groups,}
\end{array} \right.
\end{displaymath}
where we let the Hamming distances of $j$-th substring between the examples in $P^{(j)}(i))$ from 0 to be 0,1,2 to rebalance the items.

\subsubsection{Balanced binary codes in Feature-level}
To reduce the running time for candidate codes checking, the false positives in candidate set should be small, that is to minimize $\bigcup_{i=1}^{j} \mathcal{N}_j^{r'}(\mathbf{q})-\mathcal{R}^{r}(\mathbf{q})$. To achieve it, the $\bigcup_{i=1}^{j} \mathcal{N}_j^{r'}(\mathbf{q})$ should not contains too many items which are not true $r$-neighbors of query. That is, when the substring $\mathbf{h}^{(j)}$ and $\mathbf{q}^{(j)}$ differ by $r'$ bits, the full Hamming distant between $\mathbf{q}$ and $\mathbf{h}$ should differ by $(r'+1)m$ bits or less. This leads to the follow proposition:

\begin{proposition}
Suppose that for all $\mathbf{h}$ and $D(\mathbf{q},\mathbf{h}) = r$, we have $D(\mathbf{q}^{(i)},\mathbf{h}^{(i)}) = r'+1, \forall i=1,\cdots,a$, and $D(\mathbf{q}^{(i)},\mathbf{h}^{(i)}) = r', \forall i=a+1,\cdots,m$, where $r = m*r' + a$, then $\bigcup_{i=1}^{j} \mathcal{N}_i^{r'}(\mathbf{q})-\mathcal{R}^{r}(\mathbf{q}) = \emptyset$.
\label{proposition2}
\end{proposition}

According to Proposition~\ref{proposition2}, we add the following new balanced constraint in our objective
\begin{equation}
\begin{aligned}
& \psi(\mathbf{h},\mathbf{q}) = \sum_{j=1}^m  \max(0, r' - D(\mathbf{h}^{(j)},\mathbf{q}^{(j)})) & \\
& +\max(0, D(\mathbf{h}^{(j)},\mathbf{q}^{(j)}) - (r'+1)), &\\
\end{aligned}
\label{feature_level}
\end{equation}
where $r' = \lfloor D(\mathbf{h},\mathbf{q}) / m \rfloor$. The above formulation requires the almost equal distance in each substring, which distance of each substring should be less or equal to $r'+1$ and larger or equal to $r'$.

Overall, the similarity-preserving loss function for balanced codes can be formulated as :
\begin{equation}
\begin{aligned}
& \min \sum_{i,i^+,i^-} \max(0, \epsilon + D(\mathbf{h}_i,\mathbf{h}_{i^+}) -  D(\mathbf{h}_i,\mathbf{h}_{i^-})),& \\
& + \lambda \sum_{i,j} \psi(\mathbf{h}_i,\mathbf{h}_j) + \beta \sum_{i,j} \phi(P^{(j)}(i)) &
\end{aligned}
\label{objective}
\end{equation}
where $\epsilon,\lambda,\beta$ are parameters, $D(\mathbf{h}_i,\mathbf{h}_j)$ is the distance between two binary codes. For ease of optimization, we replace the Hamming distance with the euclidean distance. In all our experiments, the $\epsilon$ is setted to be $l/2$,  $\lambda=\beta = 0.1$, and $k'=20$. The first term of the objective is to preserves relative similarities of the form ``$\mathbf{h}_i$ is more similar to $\mathbf{h}_{i^+}$ than to $\mathbf{h}_{i^-}$''. The second term is for generating balanced codes in feature-level and the third term is for balanced codes in instant-level.

\begin{table*}[t]
\small
    \centering \caption{Speed-up factors for MIH vs. linear scan.}
    \begin{tabular}{|c|c | c | c |c|c | c | c | c| c|}
        \hline
 Dataset & nbits & Method & 1-NN & 2-NN & 5-NN & 10-NN & 20-NN &  50-NN & 100-NN \\
        \hline
   \multirow{8}{*}{ NUS-WIDE } &  \multirow{2}{*}{ 64 } & DMIH & \bf{119.23} & \bf{100.37} &\bf{ 87.62} & \bf{77.59} & \bf{ 66.75} & \bf{50.01} & \bf{34.68}  \\
   & & DeepHash & 56.17 & 55.28 & 49.34 & 43.79 & 36.82 & 30.70 & 22.71\\
      \cline{2-10}
    &    \multirow{2}{*}{ 96 } & DMIH & \bf{74.59} & \bf{60.51} & \bf{52.78} & \bf{44.02} & \bf{41.93} & \bf{30.96} & \bf{26.08}  \\
   & & DeepHash & 40.83 & 34.25 & 33.08 & 20.06 & 21.18 & 16.61 & 12.08\\
      \cline{2-10}
    &    \multirow{2}{*}{ 128 } & DMIH & \bf{60.90} & \bf{47.66} & \bf{38.79} & \bf{35.56} & \bf{19.49} & \bf{18.85} & \bf{17.36}\\
   & & DeepHash & 30.95 & 24.00 & 21.30 & 19.97 & 11.82 & 10.11 & 9.37\\
      \cline{2-10}
       &    \multirow{2}{*}{ 256 } & DMIH & \bf{30.31} & \bf{23.91} & \bf{22.28} & \bf{20.62} & \bf{20.72} & \bf{16.21} & \bf{14.69}\\
   & & DeepHash & 15.24 & 12.75 &  12.45 & 13.42 & 12.45 & 9.51 & 8.62\\
      \hline
   \multirow{8}{*}{ SVHN } &  \multirow{2}{*}{ 64 } & DMIH & \bf{96.39} & \bf{63.86} & \bf{65.74} & \bf{65.76} & \bf{64.15} & \bf{43.83} & \bf{31.95}   \\
   & & DeepHash & 10.06 & 10.33 & 10.29 & 10.25 & 10.03 & 10.93 & 9.79\\
      \cline{2-10}
    &    \multirow{2}{*}{ 96 } & DMIH & \bf{92.46} & \bf{64.02} & \bf{58.08} & \bf{54.09} & \bf{45.82} & \bf{25.48} & \bf{20.44}   \\
   & & DeepHash & 9.96 & 10.01 & 9.79 & 10.24 & 9.71 & 9.06 & 10.06 \\
      \cline{2-10}
    &    \multirow{2}{*}{ 128 } & DMIH & \bf{56.82} & \bf{49.35} & \bf{22.17} & \bf{18.33} & \bf{16.33 }& \bf{17.94} & \bf{17.59} \\
   & & DeepHash & 9.12 & 9.34 & 7.56 & 6.21 &  6.16 & 6.97 & 6.39\\
      \cline{2-10}
       &    \multirow{2}{*}{ 256 } & DMIH & \bf{30.27} & \bf{23.94} & \bf{21.14} & \bf{19.46} & \bf{17.91} & \bf{ 14.87} & \bf{13.19} \\
   & & DeepHash & 7.53 & 7.60 & 7.02 & 7.44 & 7.62 & 7.20 & 7.34 \\
   \hline
        \end{tabular}
    \label{speed}
\end{table*}

\section{Experiments}
In this section, we evaluate and compare the performance of the proposed method with several state-of-the-art algorithms.

\subsection{Datasets and Experimental Setting}

\begin{itemize}
%\item \textbf{ImageNet~\cite{ILSVRC15}~\footnote{http://image-net.org/challenges/LSVRC/2012/}} is a large hand-labeled image dataset, which consists of about 1.2 million images and 1,000 categories. 

\item \textbf{SVHN~\cite{svhn} \footnote{http://ufldl.stanford.edu/housenumbers/}} is obtained from house numbers in Google Street View images, which contains over 600,000 images and 10 classes.

\item \textbf{NUS-WIDE~\cite{nus-wide-civr09}~\footnote{http://lms.comp.nus.edu.sg/research/NUS-WIDE.htm}} consists of 269,648 images and the associated tags from Flickr. The labels extracted from the tags associated to the images for the 81 concepts.

\end{itemize}

%In ImageNet, we choose the images from the first 100 classes as the retrieval database,
%~\footnote{Due to the time-consuming DNN training process, e.g., training whole ImageNet dataset by AlexNet model needs several days to achieve a good hash model, we thus use a subset of ImageNet to faster evaluate the proposed method. }
%which contains more than 100,000 images. And all 5,000 images (50 images pre class) of the first 100 classes from its validation set are selected as the test query set. We uniformly sample 500 images per class from the retrieval database to form the training set.

In NUS-WIDE, we follow the settings in~\cite{CNNH,AGH} for fair comparison. The 21 most frequent labels are selected, where each label associates with at least 5,000 images. We randomly select 100 images from each of the selected 21 classes to form the query set of 2,100 images. The rest images are used as the retrieval database. In the retrieval database, 500 images from each of the selected 21 classes are randomly chosen as the training set.

In SVHN, we randomly select 1,000 images (100 images per class) as the query set, and 5,000 images (500 images per class) from the rest images as the training set. 

%100 images per class are selected as the test query set in SVHN and NUS-WIDE, i.e., 1,000 images (10 classes) and 2,100 images (21 classes), respectively. Other images are used as the retrieval database.  In all datasets, we resize images of all these databases into $256\times 256$. And 500 images per class from the retrieval databases are randomly chosen as the training set.

We implement the proposed method using the open-source \textit{Caffe}~\cite{jia2013caffe} framework. In this paper, we use AlexNet~\cite{AlexNet} as our basic network. The weights of layers are firstly initialized by the pre-trained AlexNet model~\footnote{http://dl.caffe.berkeleyvision.org/bvlc\_alexnet.caffemodel}.

\subsection{Results}
In this subsection, we evaluate the query time of our method by comparing it with the existing deep-network-based method. To make a fair comparison, we compare two methods:
\begin{itemize}
\item \textbf{DeepHash}. The hash functions are learned without the assistance of the balanced constraints, i.e., only use the first term of the objective (\ref{objective}).

\item \textbf{Deep Multi-Index Hashing (DMIH)}. The hash functions are learned with the assistance of the balanced constraints, i.e., using all terms in the objective (\ref{objective}).
\end{itemize}

Since the two methods use the same network and the only difference is that using or not using the proposed balanced constraints in feature-level and instance-level, these comparisons can show us whether the balanced constraints can contribute to the speed or not.

After obtaining the binary codes, we use the implementation of MIH~\footnote{https://github.com/norouzi/mih} provided by the authors to report the accelerated ratios of the two methods compared to linear scan on all the above databases. The speed-up factors of MIH over linear scan of both the proposed method and DeepHash for different $k$NN problems are shown in Table~\ref{speed}. Note that the linear scan does not depend on the underlying distribution of the binary codes, thus the running times of linear scan of two methods are the same.

The results show that DMIH is more efficient than DeepHash, especially for the small $k$NN problems. For instance, for $1$-NN in 96-bits codes on SVHN, the speed-up factor for DMIH is 92.46, compared to 9.96 for DeepHash. In NUS-WIDE, our method shows about $2$ speed-up ratio in comparison with DeepHash. 
%On ImageNet, the speed-up ratio of our method is ? for $10$-NN in 96-bits codes, compared to ? of DeepHash.

The main reason is that the proposed method can learn the more balanced hash codes than DeepHash. To give an intuitive understanding of our method, we utilize an entropy based measurement that is defined as
\begin{equation}
E(H) = -\frac{1}{m} \sum_{j=1}^{m} \sum_{i=1}^{2^{l_j}} p(i) \times \log(p(i)),
\end{equation}
where $l_j = l/m$ is the dimension of the $j$-th hash table, thus there are $2^{l_j}$ buckets in this table. And $p(i)$ is the probability of codes assigned to bucket $i$, which is defined as  $p(i) = n(i)/N$, where $n(i)$ is the number of codes in bucket $i$ and $N$ is the size of database. Note that the higher entropy value means better distribution of data items in hash tables. 

\begin{table}[h]
\small
    \centering \caption{Distribution of data items in the hash tables of the two methods.}
    \begin{tabular}{|c|c c c c|}
        \hline
 Method  & 64 bits & 96 bits & 128 bits & 256 bits \\
        \hline
        \multicolumn{5}{|c|}{SVHN} \\
        \hline
        DeepHash & 4.06 & 4.32 & 4.14 & 4.11 \\
         \hline
           DMIH & \bf{10.40} & \bf{10.94} & \bf{9.10} &\bf{ 8.17}\\
          \hline
           \multicolumn{5}{|c|}{NUS-WIDE} \\
        \hline
        DeepHash & 9.23 & 9.59 & 8.99 & 8.97 \\
         \hline
           DMIH & \bf{9.72} & \bf{9.84} &\bf{9.51} & \bf{9.39}\\
          \hline
        \end{tabular}
    \label{entropy}
\end{table}
Again, for all databases and bits, our method yields the higher entropy and beats the baseline. This is also can explain why our method can obtain the faster searching.

\begin{figure*}[ht!]
%Requires \usepackage{graphicx}
  \begin{flushleft}
  \centering
 
  {\includegraphics[width=0.32\textwidth]{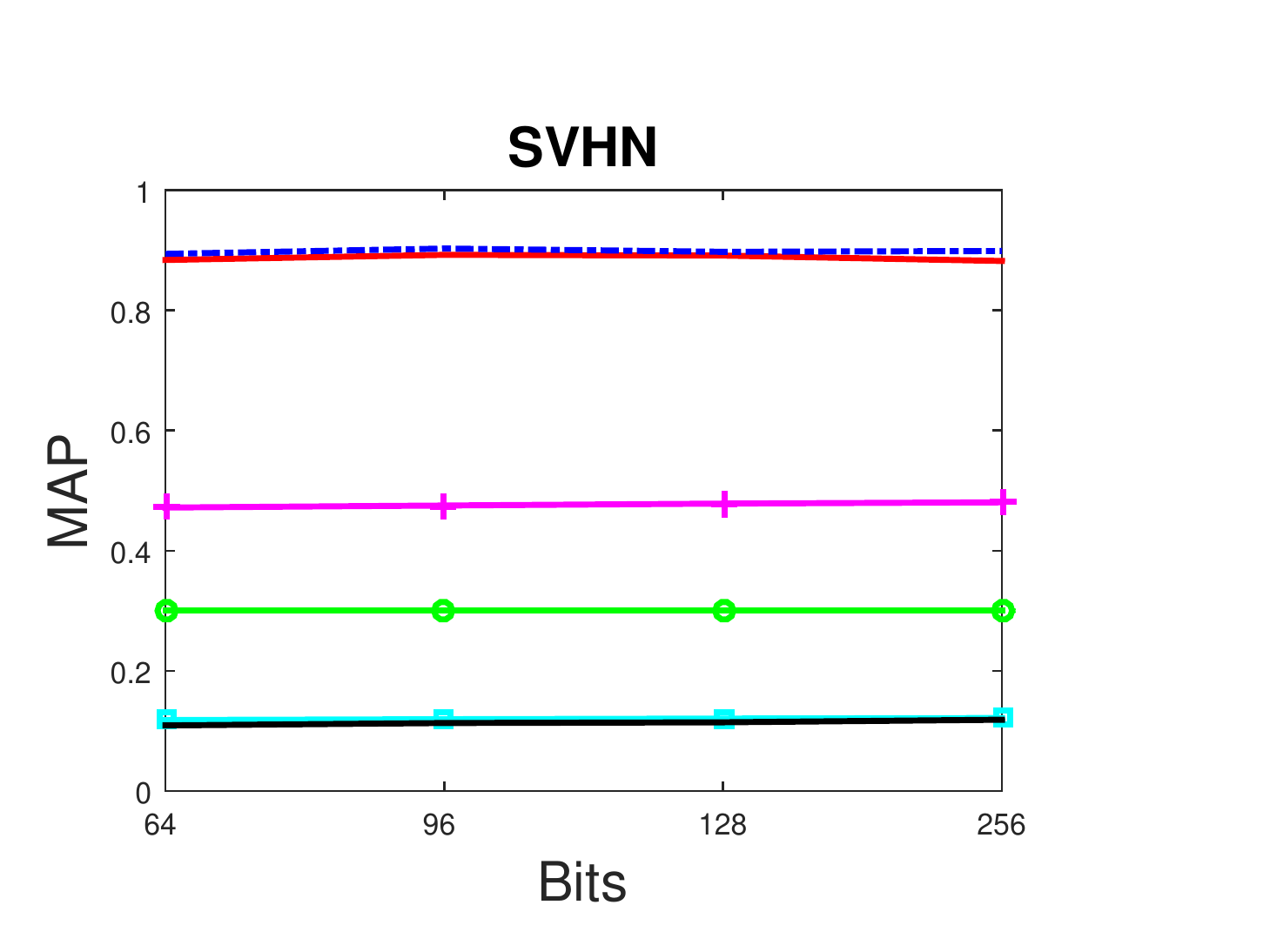}
  }
  {\includegraphics[width=0.32\textwidth]{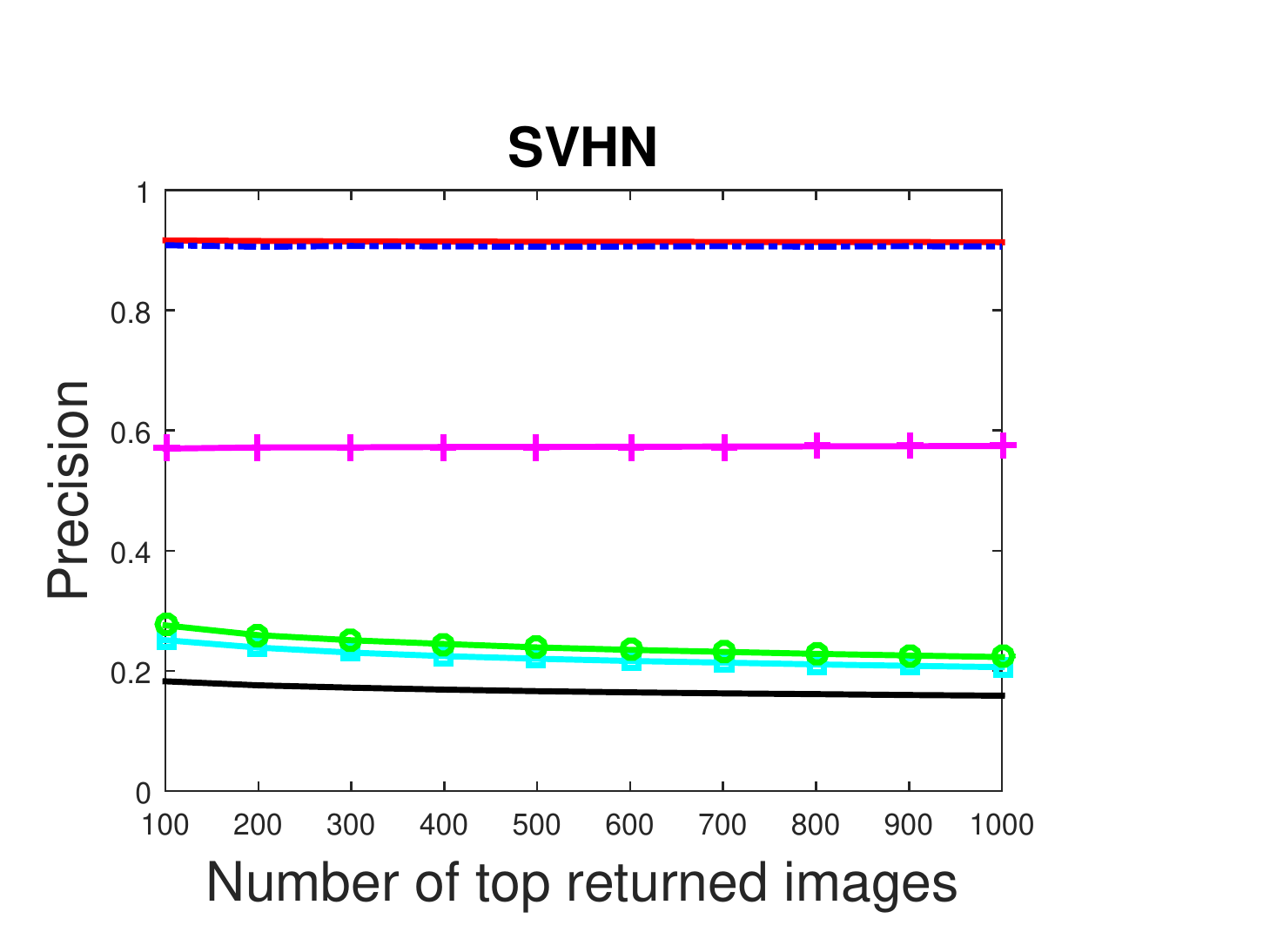}
  }
  {\includegraphics[width=0.32\textwidth]{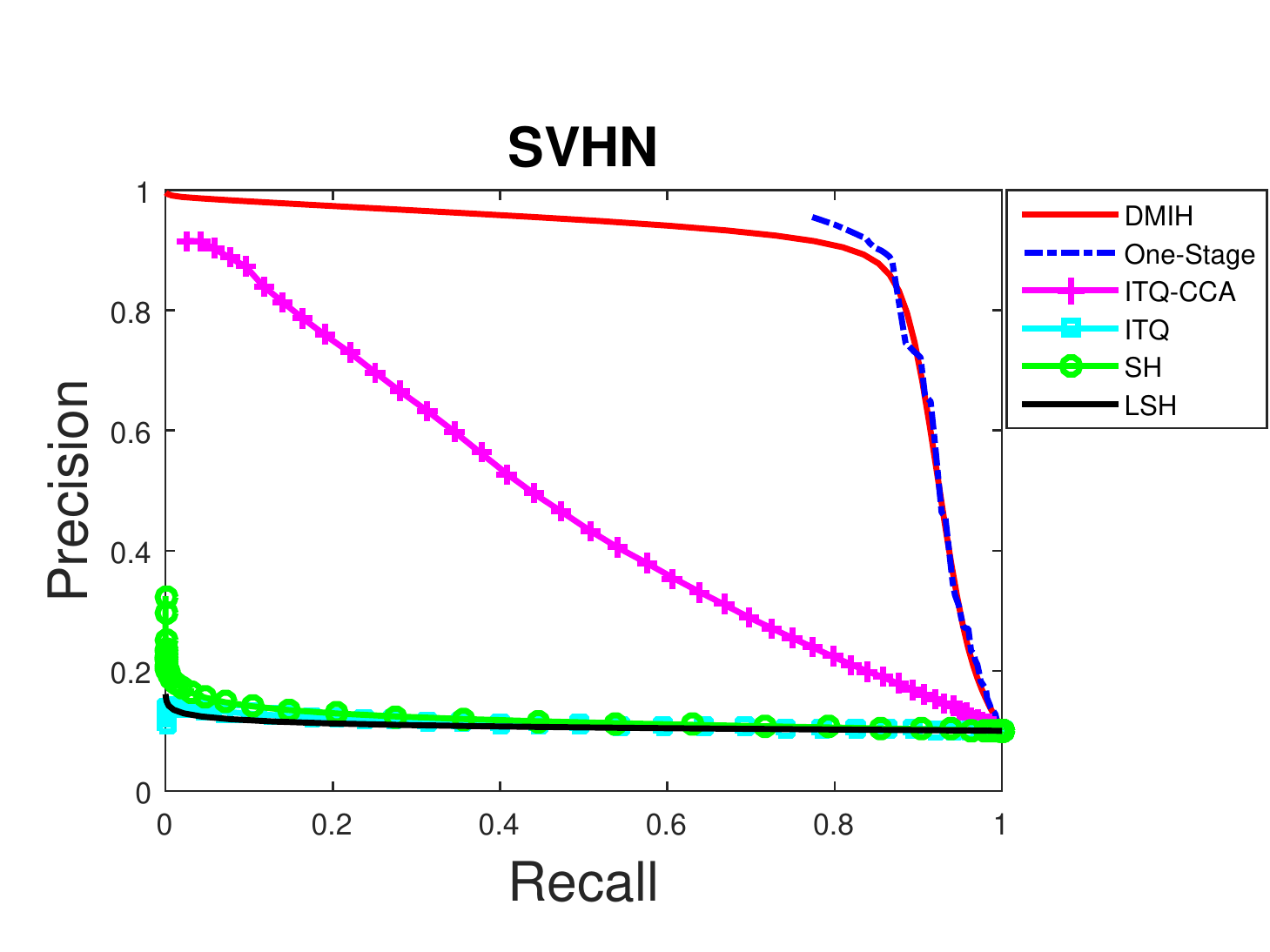}
  }
   \subfigure[]{\label{NUS-WIDE-a}
   \raisebox{-0.01cm}{\includegraphics[width=0.31\textwidth]{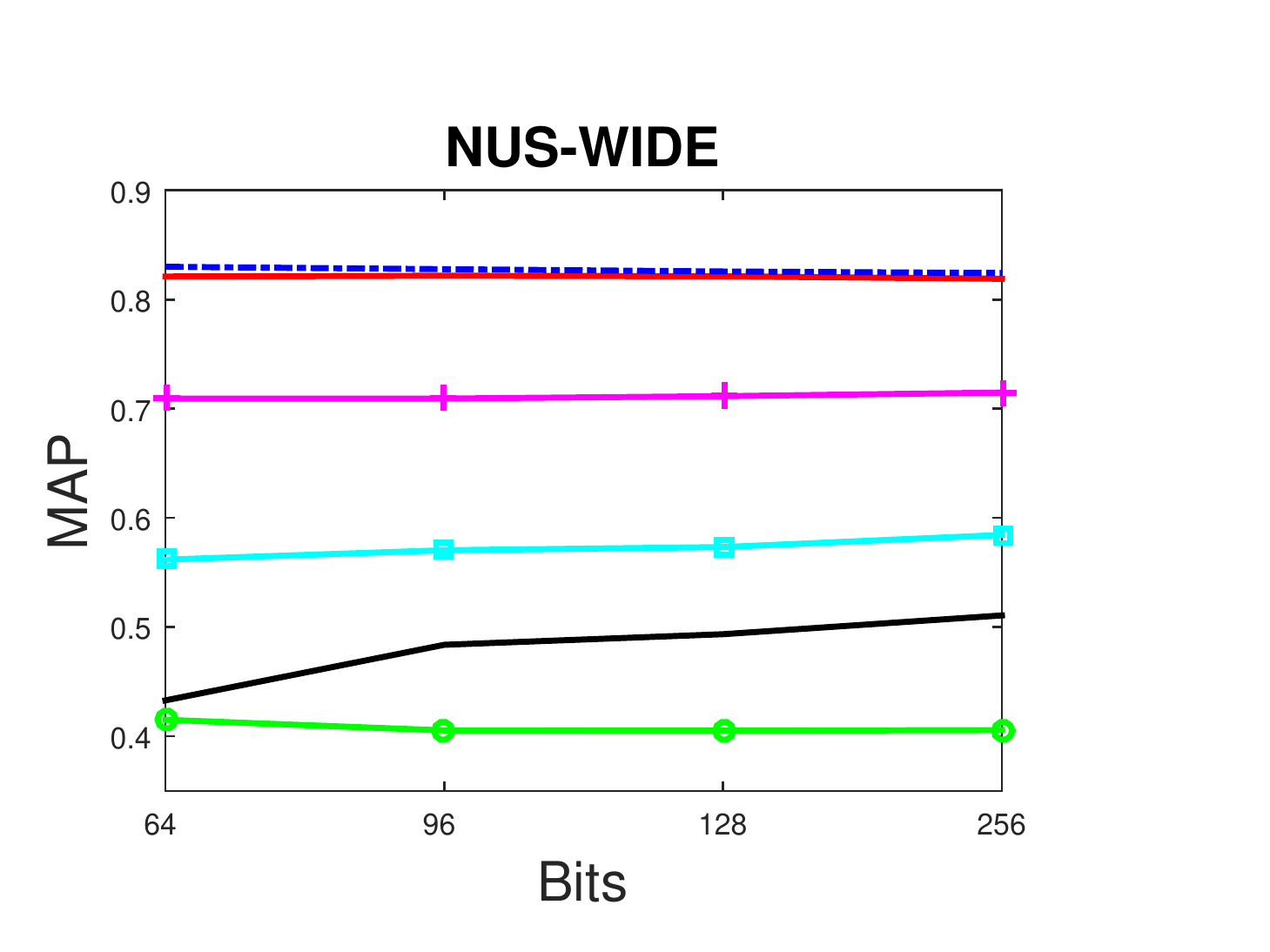}
  }}
  \subfigure[]{\label{NUS-WIDE-b}
  \raisebox{-0.01cm}{\includegraphics[width=0.31\textwidth]{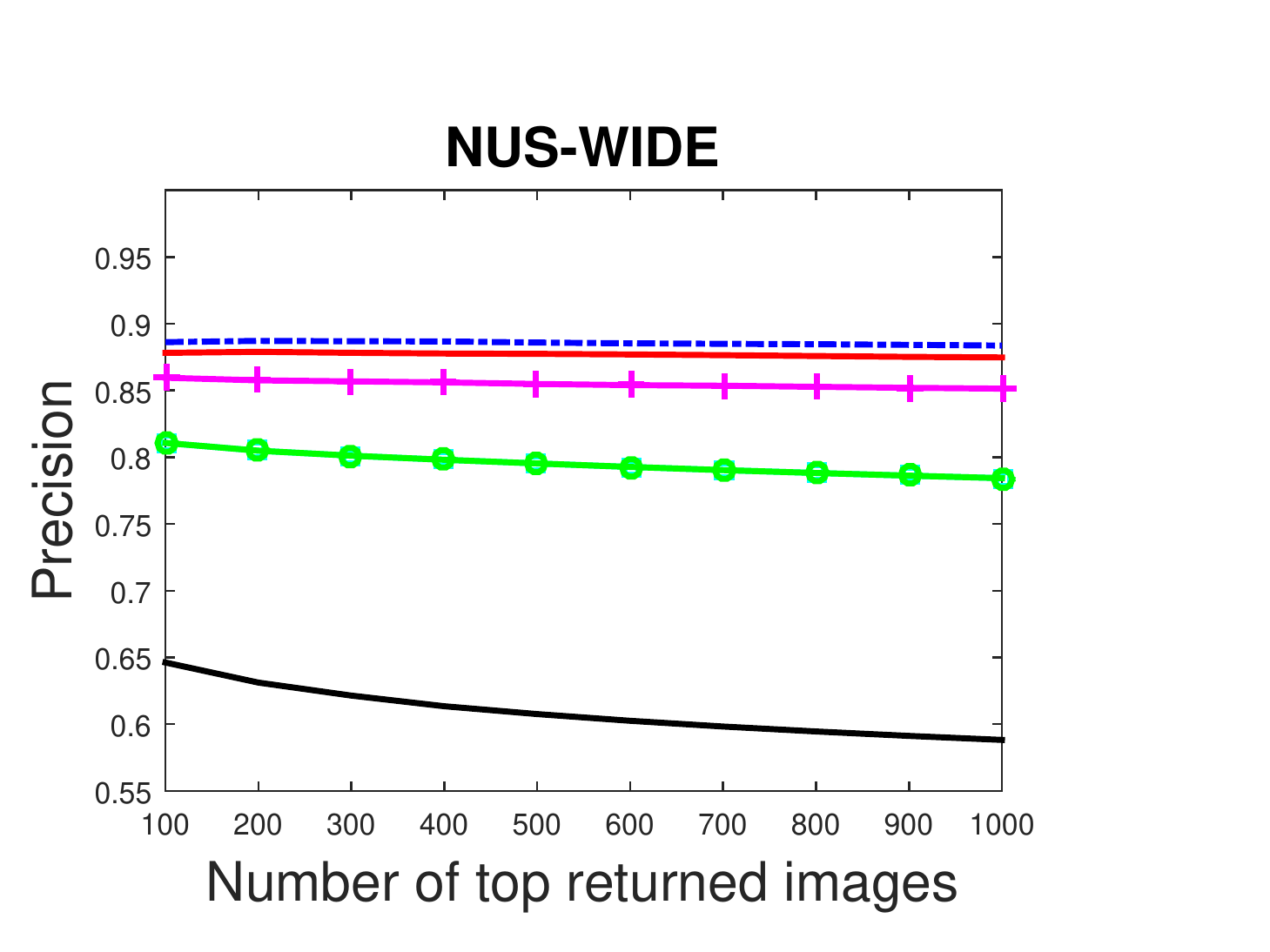}
  }}
  \subfigure[]{\label{NUS-WIDE-c}
  \raisebox{-0.01cm}{\includegraphics[width=0.31\textwidth]{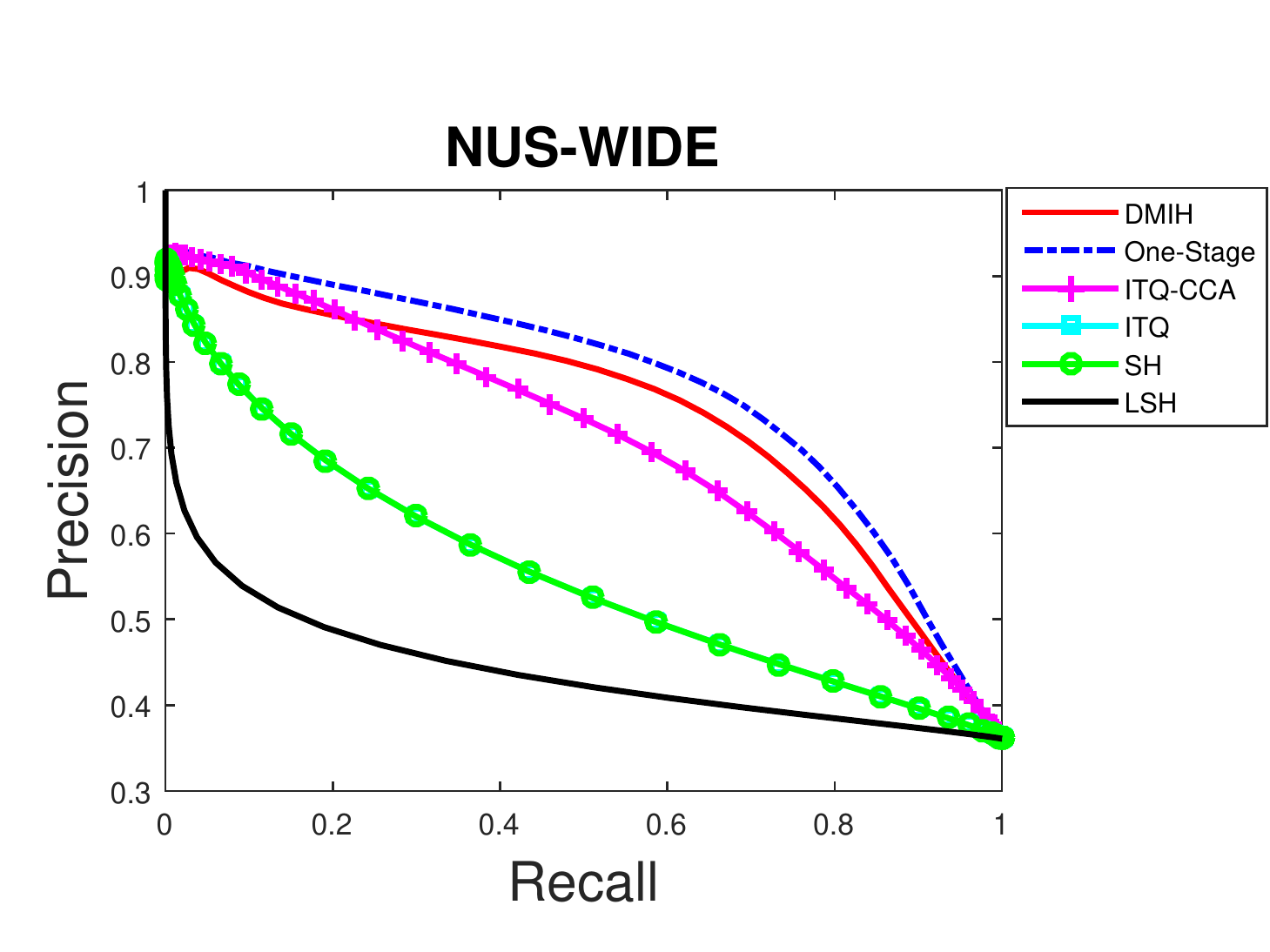}
  }}
  \caption{\footnotesize The comparison results on SVHN and NUS-WIDE. (a) MAP w.r.t. different number of bits; (b) precision curves with 64 bits w.r.t different numbers of top returned samples. (c) precision-recall curves of Hamming ranking with 64 bits; }
  \label{map-result}
  \end{flushleft}
\end{figure*}

Further, we evaluate and compare the performance of the proposed method with several state-of-the-art algorithms. LSH~\cite{LSH},
ITQ~\cite{ITQ}, ITQ-CCA~\cite{ITQ}, SH~\cite{sh} and DeepHash are selected as the baselines. The results of LSH, ITQ, ITQ-CCA and SH are obtained by the implementations provided by their authors, respectively. Note that DeepHash is very similar to the existing work One-Stage Hash~\cite{onestep}, which also divides the feature into several slices and uses the triplet ranking loss for preserving the similarities. Since the results of DeepHash and One-stage are almost the same, thus we only report the results of DeepHash. To evaluate the quality of hashing, we use Mean Average Precision ({MAP}) and Precision curves w.r.t. different numbers of top returned samples as the evaluation metrics. For a fair comparison, all of the methods use identical training and test sets, and the AlexNet model is used to extract deep features (i.e., 4096 dimensional features from fc7 layer) for LSH, ITQ, ITQ-CCA and SH. 

Figure~\ref{map-result} shows the comparison results on the two datasets. We can see that 1) the deep network based methods show improvements over the baselines using fixed deep features. 2) DMIH shows comparable performance against the most related baseline DeepHash. These results verify that adding new balanced constraints does not drop the performance. 

In summary, our method performs 2 to 10 times faster than DeepHash with the comparable performance.

\subsubsection{Effects of the Feature-level and Instance-level Constraints}
In this set of experiments, we show the advantages of the proposed two balanced constraints. To give an intuitive comparison, we show the results of using only the feature-level/instance-level constraint, respectively.  

\begin{table}[t]
\small
    \centering \caption{Speed-up factors for MIH vs. linear scan.}
    \begin{tabular}{|c | c |c|c |}
        \hline
  Method & 1-NN & 10-NN &  100-NN \\
        \hline
  \multicolumn{4}{|c|}{NUS-WIDE} \\
       \hline
     Both & \bf{ 74.59} & \bf{ 44.02} &\bf{26.08}\\
  Feature-level&68.00 & 37.64 & 22.29 \\
  Instance-level & 45.48 & 30.22 & 17.82 \\
%     \hline
%   Both & \\
%  & Feature & \\
%  & Instance & \\
     \hline
        \multicolumn{4}{|c|}{SVHN} \\
     \hline
         Both & \bf{92.46} & \bf{54.09} & \bf{20.44}  \\
  Feature-level & 11.73 & 10.43 & 9.99 \\
  Instance-level & 89.73 & 50.23 & 19.94 \\
     \hline
%         \multirow{3}{*}{ 96 } & Both & \\
%  & Feature & \\
%  & Instance & \\
%   \hline
        \end{tabular}
    \label{speed_2}
\end{table} 

Table~\ref{speed_2} show the comparison results. The results show that instance-level constraint is very useful for SVHN while the feature-level constraint is helpful in NUS-WIDE dataset. It depends on the data distributions of the learned binary codes.

\subsubsection{Effect of the End-to-end Learning}
Our framework is an end-to-end framework. To show the advantages of the end-to-end framework, we compare to the following baseline, which adopts a two-stage strategy. In the first stage,  DeepHash is learned and the images are encoded into binary codes. In the second stage, we rebalance the binary codes by the data-driven multi-index hashing~\cite{wan2013data}.

\begin{table}[t]
\small
    \centering \caption{Speed-up factors for MIH vs. linear scan.}
    \begin{tabular}{|  c | c |c|c |}
        \hline
 Method & 1-NN & 10-NN &  100-NN \\
        \hline
  \multicolumn{4}{|c|}{NUS-WIDE} \\
       \hline
    DMIH & \bf{ 74.58} & \bf{ 44.02} & \bf{26.08} \\
  DeepHash & 40.83 & 20.06 & 12.08\\
  Two-stage & 60.55 & 29.97 & 17.95 \\
     \hline
        \multicolumn{4}{|c|}{SVHN} \\
     \hline
     DMIH & \bf{92.46} & \bf{ 54.09} & \bf{20.44} \\
  DeepHash & 9.96 & 10.24 & 10.06 \\
  Two-stage & 11.50 & 10.40 & 10.33 \\
   \hline
        \end{tabular}
    \label{speed_3}
\end{table} 
Table~\ref{speed_3} shows the comparison results. We can observe that our method performs better than DeepHash and two-stage method. It is desirable to learn the hash function and balanced procedure in the end-to-end framework.

\section{Conclusion}
\label{conclusion}
In this paper, we proposed a deep-network-based multi-index hashing method for fast searching and good performance. In the proposed deep architecture, an image goes through the deep network with stacked convolutional layers and is encoded into high level image representation with several substrings. Then, we proposed to learn more balanced binary codes by adding two  constraints. One is the feature-level constraint, which is used to make the binary codes distributed as balance as possible in each hash table. Another is the instance-level constraint, which is used to let the buckets in each substrings hash table contain balanced items. Finally, the deep hash model for both the powerful image representation and fast searing is learned simultaneously. Empirical evaluations on two datasets show that the proposed method runs faster than the baseline and achieve comparable performance. 

In future work, we plan to apply DMIH in different networks and methods to exploit the effect of the proposed balanced constraints.  We also plan to accelerate the running times of extracting the features from the deep network.

\section*{Appendices}
\subsection*{Proof of Proposition 1}
Suppose that there exists one binary code $\mathbf{h}$, $\mathbf{h} \in \mathcal{R}^{r}(\mathbf{q})$ and  $\mathbf{h} \notin \bigcup_{i=1}^{j} \mathcal{N}_j^{r'}(\mathbf{q})$. According to Proposition 1 in paper~\cite{mih}, we have $D(\mathbf{h}^{(z)},\mathbf{q}^{(z)}) \leq r'$ for a substring. We discuss in two situations: 1) if $z \leq j$, then $\mathbf{h} \in \bigcup_{i=1}^{j} \mathcal{N}_i^{r'}(\mathbf{q})$ according to the definition of $\mathcal{N}_z^{r'}(\mathbf{q})$, which contradicts the premise. 2) if $z > j$ and all the first $j$ substrings is strictly greater than $r'$, then the total number of bits that differ in the last $m-j$ is at most $r - j*(r'+1) = r'm+j-1 -jr'-j = (m-j)r'-1$. Using Proposition 1 in paper~\cite{mih} again, we have $D(\mathbf{h}^{(z)},\mathbf{q}^{(z)}) \leq r'-1$, thus $\mathbf{h} \in \mathcal{N}_z^{r'-1}(\mathbf{q}) \subseteq \mathcal{N}_j^{r'}(\mathbf{q})$, which contradicts the premise.

\subsection*{Proof of Proposition 2}
Suppose that $\bigcup_{i=1}^{j} \mathcal{N}_i^{r'}(\mathbf{q})-\mathcal{R}^{r}(\mathbf{q}) \neq \emptyset$, we have at least one binary code $\mathbf{b}$ satisfies $\mathbf{b} \in \bigcup_{i=1}^{j} \mathcal{N}_i^{r'}(\mathbf{q})$ and $\mathbf{b} \notin \mathcal{R}^{r}(\mathbf{q})$. Since $\mathbf{b}$ is not the $r$-neighbor of query $\mathbf{q}$, then $\mathbf{b}$ and $\mathbf{q}$ at least differ by $r+1$ bits. Since $\mathbf{b} \in \bigcup_{i=1}^{j} \mathcal{N}_i^{r'}(\mathbf{q})$, we have $r = r'm + j -1 = r'm + a$, thus $j=a+1$. According to the assumption and $D(\mathbf{q},\mathbf{h}) = r+1$, we have $D(\mathbf{q}^{(j)},\mathbf{h}^{(j)}) = r'+1$,  thus $\mathbf{b} \notin \bigcup_{i=1}^{j} \mathcal{N}_i^{r'}(\mathbf{q})$ by the definition of $\mathcal{N}_j^{r'}(\mathbf{q})$ , which contradicts the premise.

%\begin{table*}[ht!]
%\small
%    \centering \caption{Comparison of shared sub-network against multiple independent sub-networks on the three databases.}
%    \begin{tabular}{|c|c c c c|c c c c|c c c c|}
%         \hline
%        \multirow{2}{*}{ Method } & \multicolumn{4}{|c}{SVHN(MAP)} &\multicolumn{4}{|c}{CIFAR-10(MAP)} & \multicolumn{4}{|c|}{NUS-WIDE(MAP)}\\
%& 12 bits & 24 bits & 32 bits & 48 bits & 12 bits & 24 bits & 32 bits& 48bits & 12 bits & 24 bits &32 bits & 48 bits \\
%        \hline
%        shared  & 0.894 & 0.914 & 0.925 & 0.923 & 0.552 & 0.566 & 0.558 & 0.581  &  0.674 & 0.697 & 0.713 & 0.715 \\
%         \hline
%        independent & &  \\
%         \hline
%        \end{tabular}
%    \label{sub-network}
%\end{table*}

{\small
\bibliographystyle{ieee}
\bibliography{0812}
}

\end{document}